# Hybrid Dense-UNet201 Optimization for Pap Smear Image Segmentation Using Spider Monkey Optimization

Ach Khozaimi[1,2], Isnani Darti[1], Syaiful Anam[1], Wuryansari Muharini Kusumawinahyu[1]
[1] Department of Mathematics, Faculty of Mathematics and Natural Sciences, Brawijaya University, Malang, Indonesia
[2] Department of Computer Science, Faculty of Engineering, Universitas Trunojoyo Madura, Indonesia

**ABSTRACT** (10 PT)

Pap smear image segmentation is crucial for cervical cancer diagnosis. However, traditional segmentation models often struggle with complex cellular structures and variations in pap smear images. This study proposes a hybrid Dense-UNet201 optimization approach that integrates a pretrained DenseNet201 as the encoder for the U-Net architecture and optimizes it using the spider monkey optimization (SMO) algorithm. The Dense-UNet201 model excelled at feature extraction. The SMO was modified to handle categorical and discrete parameters. The SIPaKMeD dataset was used in this study and evaluated using key performance metrics, including loss, accuracy, Intersection over Union (IoU), and Dice coefficient. The experimental results showed that Dense-UNet201 outperformed U-Net, Res-UNet50, and Efficient-UNetB0. SMO Dense-UNet201 achieved a segmentation accuracy of 96.16%, an IoU of 91.63%, and a Dice coefficient score of 95.63%. These findings underscore the effectiveness of image preprocessing, pretrained models, and metaheuristic optimization in improving medical image analysis and provide new insights into cervical cell segmentation methods.

Ach Khozaimi
Department of Computer Science, Faculty of Engineering, Universitas Trunojoyo Madura, Indonesia
Email: khozaimi@trunojoyo.ac.id

## 1. INTRODUCTION

Cancer is the second leading cause of death worldwide, after heart disease [1]. Cervical cancer is a major health concern for women worldwide. Cervical cancer cells can spread to other organs [2]. The Global Cancer Observatory (GCO) reported 570,000 new cases and 311,000 deaths due to cervical cancer in 2018 [3]. In Indonesia, it is the second most common cancer after breast cancer, with 36,633 cases reported in 2020 [4]. By 2040, cervical cancer is expected to cause 460,000 deaths annually, with 90% of these deaths occurring in developing countries [5]. Early screening, such as Pap smears (Papanicolaou smears), is crucial for reducing the number of cases [6]. Pap smear screening plays a vital role in the early detection and classification of cervical cancer. Because it relies on medical images, computer-aided analysis enhances accuracy. Pap smear image analysis generally involves five stages: preprocessing, segmentation, feature extraction, feature selection, and classification, followed by performance evaluation [7]. Optimizing the parameters or hybridizing methods at each stage can further improve model performance [8]. However, the manual examination of Pap smear slides is time-consuming, subject to inter-observer variability, and prone to human error. To address these challenges, automated image segmentation techniques have been developed to improve the accuracy and efficiency of cervical cell analysis [9].

Semantic segmentation models, such as U-Net and its variants, have shown promising results in medical image segmentation [10]. U-Net's ability to capture fine-grained spatial details makes it suitable for segmenting complex cellular structures in medical images [11]. However, standard U-Net architectures may struggle with variations in cell morphology, overlapping regions, and noise in microscopic images, which limits their segmentation accuracy [12]. To overcome these limitations, many studies have replaced the U-Net encoder with more advanced CNN architectures [13]. CNN architecture enhances feature extraction, improves training efficiency, and leverages pretrained weights for transfer learning [14]. The original U-Net encoder consists of standard convolutional and max-pooling layers; however, architectures such as ResNet, VGG, and EfficientNet have been explored as alternatives. Res-UNet benefits from residual connections that help prevent vanishing gradients[15]. VGG-UNets provide a simple yet effective feature extractor [16], and Efficient-UNet achieves high accuracy with fewer parameters[17]. However, other CNN architectures must be explored to improve U-Net performance.

In this study, we propose a hybrid Dense-UNet201 optimization approach for the segmentation of pap smear images. The pretrained DenseNet201 was used as the encoder in the U-Net model. Many studies have shown that DenseNet201 has demonstrated outstanding performance in medical image classification. DenseNet201 outperformed InceptionResNetV2, VGG19, and Xception in cervical cancer classification using the SIPaKMeD dataset [18]. DenseNet-201outperformed several deep learning architectures, including VGG-16, VGG-19, ResNet-101, ResNet-152, DenseNet-121, Xception, DenseNet-169, MobileNet, ResNet-50, Inception, and MobileNet-v2 in Pap smear image classification [19]. DenseNe201 also shown superior performance over ResNet-152, CheXNet, Xception, VGG-19, and MobileNetV2 when handling imbalanced medical image datasets [20]. Additionally, DenseNet-201 achieved 97% accuracy in brain tumour classification [21]. A hybrid DenseNet-201 and InceptionV3 achieved an impressive accuracy of 96.54% in cervical cancer classification [22].

In contrast, the proposed method (Dense-UNet201) was optimized using spider monkey optimization (SMO). SMO is a nature-inspired optimization technique that mimics the foraging and exploratory behaviour of spider monkey swarms [23]. In this study, the SMO was modified to handle discrete and categorical parameters, such as epoch, optimizer, and batch size. Several studies have demonstrated the success of SMO in enhancing model performance. SMO has been shown to enhance the performance of intrusion detection systems in IoT environments [24]. Additionally, previous research has demonstrated that SMO outperforms several optimization algorithms, including Particle Swarm Optimization (PSO), Whale Optimization Algorithm (WOA), Gaussian Quantum-Behaved Particle Swarm Optimization (GQPSO), Cuckoo Search (CS), Bat Algorithm (BA), Differential Evolution (DE), Ant Lion Optimizer (ALO), Firefly Algorithm (FA), Grey Wolf Optimizer (GWO), Arithmetic Optimization Algorithm (AOA), Harmony Search (HS), and Dragonfly Algorithm (DA) [25]. Furthermore, SMO has been successfully applied to optimize hyperparameters in various CNN architectures [26] and deep neural networks (DNNs) [27].

This simulation used the SIPaKMeD dataset and evaluated it using several metrics, such as loss, accuracy, intersection over union (IoU), and dice coefficient. The SMO Dense-UNet201 underscores the effectiveness of image preprocessing and metaheuristic optimization in improving medical image analysis and provides new insights into cervical cell segmentation.

## 2. MATERIALS AND METHODS
### 2.1. Pap Smear Images, Preprocessing, and Annotation Process

The SIPaKMeD dataset was used to evaluate SMO Dense-UNet201. The SIPaKMeD dataset is well-structured for cervical cell analysis, including medical image classification and semantic image segmentation. It contains 4,049 isolated cervical cell images categorized into five distinct classes: superficial squamous epithelial cells (SSEC), intermediate squamous epithelial cells (ISEC), columnar epithelial cells (CEC), low-grade squamous intraepithelial lesions (LSIL), and high-grade squamous intraepithelial lesions (HSIL). Each class contained approximately 800 images. Images were acquired using microscopy imaging techniques. The SIPaKMeD dataset includes pre-segmented (masks) single-cell images that can be used for semantic segmentation [28]. The preprocessing pipeline began with image resizing and normalization. This step is crucial for maintaining consistent input dimensions for semantic segmentation models [29]. The hybrid Perona–Malik diffusion (PMD) filter and contrast-limited adaptive histogram equalization (CLAHE) were used to reduce noise and improve image contrast. The Hybrid PMD-CLAHE was optimized using the SMO algorithm. The Perona-Malik Diffusion (PMD) filter is an anisotropic diffusion technique that smooths an image while preserving important edges [30]. CLAHE is effective in increasing the visibility of minute features in images with uneven illumination, such as Pap smear images and improved VGG16, InceptionV3, and EfficientNet models [31]

The annotation process for the SIPaKMeD dataset involves several steps to ensure that the images and their corresponding ground-truth masks are properly prepared for segmentation tasks. First, all BMP image files were resized to a uniform resolution of 256 × 256 pixels and saved as square JPG files to ensure consistency in the input size. Next, the corresponding 256 × 256 square mask JPG files were created based on the information from *_cyt*. dat files, which contain the annotations for the cervical cell regions. These masks served as the ground truth for the segmentation task, clearly outlining the areas of interest (nucleus and cytoplasm) and separating them from the background [28]. Data augmentation techniques were applied to both the image and mask files to enhance the dataset further and improve the model's performance.

### 2.2. Dense-UNet201 Implementation

This study proposes a hybrid architecture named Dense-UNet201, which integrates DenseNet-201 as the encoder for the U-Net architecture to perform semantic segmentation on single cells. The Dense-UNet201 model replaces the traditional U-Net encoder with DenseNet201, leveraging its dense connectivity and feature reuse properties to improve the feature extraction and gradient flow. DenseNet-201 consists of multiple Dense Blocks that use feature concatenation to enhance the information flow across layers. Each Dense Block progressively reduces the spatial resolution while increasing the depth of the feature maps. The output sizes from DenseNet-201 must align with the U-Net decoder to ensure seamless reconstruction. The specific encoder and decoder are listed in Table 1.

Table 1. Output Size in Each Block

| Block | DenseNet-201 (Encoder) | U-Net (Decoder) |
| --- | --- | --- |
| Block 1 | 256×256×64 | 256×256×512 |
| Block 2 | 128×128×128 | 128×128×256 |
| Block 3 | 64×64×256 | 64×64×128 |
| Block 4 | 32×32×512 | 32×32×64 |
| Middle (Bottleneck) | 16×16×1024 | 16×16×1024 |

At the bottleneck, both architectures maintain an output of $16 \times 16 \times 1024$, ensuring smooth transition between encoding and decoding. The decoder utilizes transpose convolutions to progressively upsample the feature maps back to the original resolution, whereas skip connections from the encoder help retain spatial details. Hybrid pretrained DenseNet201 and U-Net benefits from enhanced feature extraction due to the dense connections, which improve gradient flow and prevent loss of fine-grained details. The Dense-UNet201 architecture is shown in Fig. 1. The architecture consists of an encoder based on DenseNet201 (pretrained on ImageNet) and a decoder based on U-Net. The DanseNet201 and U-Net explanations can be found in Sections 2.3 and 2.4, respectively.

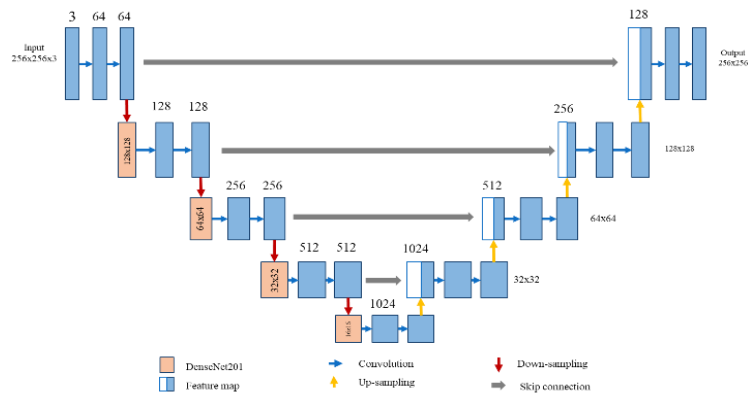

Figure 1. Dense-UNet201 Architecture

To improve the Dense-UNet201 performance, Spider Monkey Optimization (SMO) was applied to fine-tune key hyperparameters, including the learning rate (range: 1e-5 to 1e-2) as a continuous parameter, batch size (4, 8, 16, 32, 64, 128) as a categorical parameter, and number of epochs (range: 10–100) as a discrete parameter. Adamax was used for all the models. Adamax has demonstrated superior performance in Pap smear image classification [32].

The SMO algorithm was primarily designed for continuous optimization problems. However, Dense-UNet201 optimization requires handling different variable types, including discrete variables, such as the number of epochs, categorical variables (e.g., batch size), and continuous variables for the learning rate. In this study, the SMO algorithm was modified to ensure effective mixed-variable optimization. For discrete variables, such as the number of epochs (ranging from 10 to 100), a rounding mechanism was implemented to ensure that the updated values remained within valid integer ranges. After each position update in the SMO, the epoch value was rounded to the nearest valid integer. This prevents invalid fractional values.

For categorical variables, such as batch size [4, 8, 16, 32, 64, 128], a mapping strategy was used. Instead of treating the batch size as a continuous value, the SMO assigns an index to each possible batch size. During the optimization process, updates were applied to the index rather than the raw batch size value. A nearest-neighbor selection mechanism is then used to map the updated index to the closest valid batch size to ensure proper categorical handling. The update position formulas for discrete and categorical parameters, please refer to Section 2.5.

For continuous variables, such as the learning rate, the standard SMO position update mechanism remains unchanged, allowing smooth adjustments based on social learning and exploration. However, to prevent instability, an adaptive constraint is introduced to keep the learning rate within a predefined range (e.g., $10^{-5}$ to $10^{-1}$). This prevents extreme values that may lead to poor convergence of the model.

By integrating these modifications, the SMO can efficiently optimize machine learning hyperparameters that involve discrete, categorical, and continuous variables. The SMO converges when the population of solutions stabilizes or when there is no further improvement in the objective function over successive iterations. This can occur when the difference between the best solution in consecutive iterations falls below a predefined threshold, the maximum number of iterations is reached, or the global and local leaders remain unchanged for multiple iterations or are stagnant. Additionally, convergence occurs when population diversity decreases, causing all spider monkeys to cluster around a single solution, limiting further exploration.

### 2.3. DenseNet201 architecture

DenseNet201 is a deep convolutional neural network (CNN) architecture belonging to the DenseNet family, introduced by Huang et al. in 2017 [33]. DenseNet is a deep learning model that employs a feedforward approach by connecting each layer to all the subsequent layers [16]. While conventional CNN models with L layers typically have L connections, DenseNet establishes direct connections totalling $(L \times (L + 1))/2$. Each layer uses feature maps from all preceding layers as inputs, enhancing the information flow and gradient propagation. To mitigate overfitting, DenseNet incorporates regularization into the training process. In this study, we used the DenseNet-201 variant, which consists of four dense blocks containing 6, 12, 24, and 16 convolutional layers. Fig. 3 illustrates the adjusted DenseNet-201 architecture, including the classification layer [21].

The advantages of DenseNet201 include a higher classification accuracy, improved generalization, and reduced computational complexity compared with deeper networks, such as ResNet. Owing to its strong feature extraction capabilities, DenseNet201 has been widely adopted in medical image analysis, including cervical cancer classification, tumor detection, and anomaly identification.

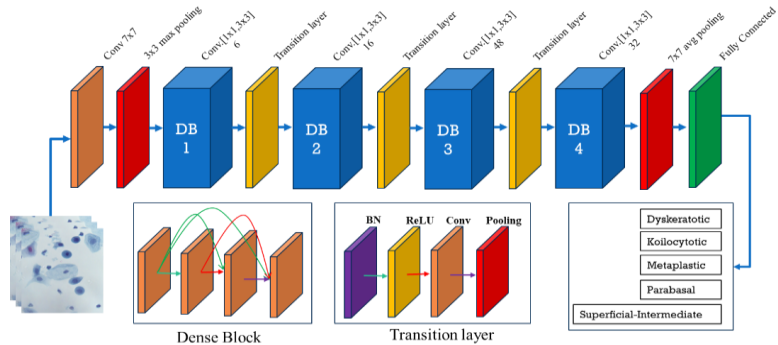

Figure 2. DenseNet201 architecture

### 2.4. U-Net Model

The U-Net architecture consists of two main paths: an encoder (downsampling path) and a decoder (upsampling path). The encoder and decoder extract features and reconstruct high-resolution segmentation maps, respectively. In the encoder path, an input image of size $572 \times 572 \times 1$ undergoes a series of $3 \times 3$ convolutions with ReLU activation, followed by $2 \times 2$ max pooling to reduce the spatial dimensions while increasing the number of filters. Each downsampling stage doubles the number of filters [64, 128, 256, and 512]. The bottleneck layer is the most abstract feature representation before the reconstruction begins with the 1024 layer.

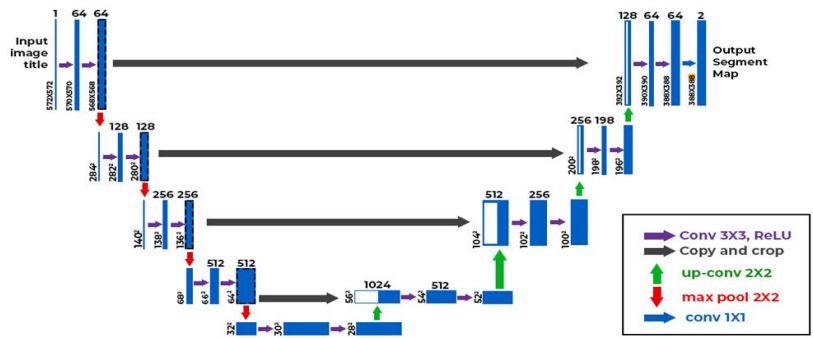

Figure 3. U-Net architecture

In the decoder path, the model employs 2 × 2 up-convolutions to enlarge the feature dimensions. copy and crop operations merge information from the encoder path to retain essential details for segmentation. This process was carried out step-by-step, reducing the number of filters from 1024 to 512, 256, 128, and finally, 64. Each upsampling stage is followed by a 3 × 3 convolution with ReLU to refine the extracted features. The final layer converts the features into an output segmentation map. With this structure, U-Net is highly effective for medical image segmentation. Fig. 4 shows the U-Net architecture [11].

**2.5. Spider Monkey Optimization**

The Spider Monkey Optimization (SMO) algorithm is inspired by the foraging behavior of spider monkeys in groups and is a population-based algorithm. It is an iterative optimization algorithm that aims to find the best solution to global optimization problems by modelling monkeys' social behavior. The SMO consists of six phases: Local Leader (*LL*), Local Leader Learning (*LLL*), Local Leader Decision (*LLD*), Global Leader (*GL*), Global Leader Learning (*GLL*), and Global Leader Decision (*GLD*) [23][34].

The process began by randomly initializing the positions of the monkeys within the problem boundaries. The position of each monkey represents a potential solution to the problem. The position of each monkey ($SM_i$) was initialised using Eq. 1.

$$SM_{ij} = SM_{minj} + R \times (SM_{maxj} - SM_{minj}) \tag{1}$$

where $R$ is a uniformly distributed random number between 0 and 1, and $SM_{min,j}$ and $SM_{max,j}$ are the lower and upper bounds for each dimension ($j$).

During the *LL* phase, monkeys update their positions based on information from their local leader and other group members. The position was updated using the following Eq. 2.

$$S_{new\ ij} = SM_{ij} + R \times (LL_{kj} - SM_{i,j}) + U(-1,1) \times (SM_{rj} - SM_{ij}) \tag{2}$$

where, $SM_{ij}$ is the $j$-th dimension of $i$-th *SM*, $LL_{kj}$ represents the $j$-th dimension of local leader of the $k$-th group and $U(-1,1)$ is a uniformly distributed random number in the range $(-1,1)$, and $SM_{rj}$ is the $j$-th dimension of a randomly selected *SM* from the $k$-th group such that $r \neq 1$.

In the GL phase, the monkeys updated their positions using information from the global leader. The update is based on the probability ($prob_i$):

$$prob_i = 0.9 \times \frac{fitness_i}{max_f itness} + 0.1 \tag{3}$$

where $fitness_i$ is the monkey's fitness and $max_f itness$ is the highest fitness among all monkeys. The position is updated as follows.

$$SM_{new\ ij} = SM_{ij} + R \times (GL_j - SM_{i,j}) + U(-1,1) \times (SM_{rj} - SM_{ij}) \tag{4}$$

where $GL_j$ is the global leader's position.

In the *GLL* and *LLL* phases, a greedy selection approach was adopted to update the leaders. If the local leader does not improve beyond a threshold (LocalLeaderLimit), the group's position is updated randomly or based on the information from the global and local leaders.

$$SM_{new\ ij} = SM_{ij} + R \times (GL_j - SM_{ij}) + U(0,1) \times (SM_{rj} - SM_{kj}) \tag{5}$$

In this study, for discrete parameters (epoch {10 – 100}), updates follow the standard SMO update equation but are rounded to the nearest integer:

$$SM_{newi} = round\ (SM_i + R \times (LL_k - SM_i) + U(-1,1) \times (SM_r - SM_i)) \tag{6}$$

where the updated value is clipped to stay within [10,100]:

$$SM_{newi} = max(10, min(100, SM_{newi})) \tag{7}$$

For categorical parameters (bith size {4, 8, 16, 32, 64, 128}), we represent values as indices (0,1,2,3,4,5) and update them probabilistically:

$$SM_{newi} = round\left(SM_i + R \times (LL_k - SM_i)\right) \qquad (8)$$

since categorical values are discrete choices, the new value is selected based on probability:

$$P(M_{newi}) = \frac{fitness(LL_k)}{\sum_j fitness(SM_j)} \qquad (9)$$

The updated categorical index is mapped back to the original category set {4, 8, 16, 32, 64, 128}. Finally, in the GLD phase, the global leader is monitored by the local leaders. If the maximum iteration limit (GlobalLeaderLimit) is reached, the group is divided into smaller subgroups for further optimization.

### 2.6. Simulation

Figure 2 shows the simulation scenario for Dense-UNet201 optimization using the SMO algorithm. It illustrates the process flow, starting with dataset loading, preprocessing, and hybrid PMD-CLAHE optimization using SMO. The SMO Dense-UNet201 model was trained using the SIPaKMeD dataset, which was improved by SMO PMD-CLAHE. The model evaluation and performance improvement steps complete the proposed workflow.

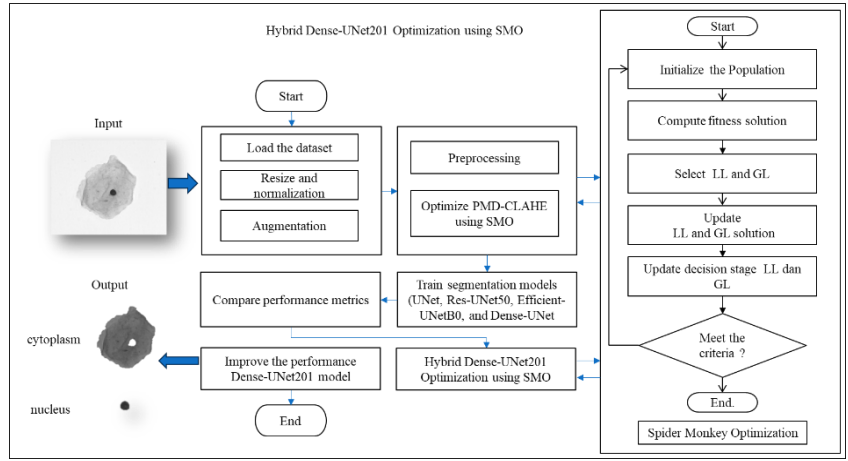

Figure 4. Dense-UNet201 model optimization using SMO

Three simulation scenarios were used in this study. First, in the initial stage, testing was conducted using the SIPaKMeD dataset without any additional processing. A hybrid Dense-U-Net201 was applied to the dataset to measure the model's baseline performance in multi-class semantic segmentation. The simulation was evaluated using loss, accuracy, IoU, and Dice coefficient. The results were compared with those of the UNet, Res-Unet50, and Efficient-UNetB0 architectures to assess the proposed method.

Second test: In the second scenario, the SIPaKMeD dataset was enhanced using a hybrid PMD filter-CLAHE. This technique aims to reduce noise and improve image contrast [35]. Subsequently, a Hybrid Dense-U-Net201 was applied for semantic segmentation. The results were compared with those of the first test to evaluate the effect of image enhancement on the model performance. In addition, comparisons were made with the UNet, Res-Unet50, and Efficient-UNetB0 architectures to determine the benefits of image quality enhancement across different models.

Third test: In the final stage, testing was conducted by optimizing Hybrid Dense-U-Net201 using the SMO algorithm. This optimization aims to fine-tune the model parameters for the best segmentation performance. This process was carried out on SIPaKMeD with a hybrid PMD filter-CLAHE enhancement. The segmentation results were then compared with those of the second simulation to assess the SMO optimization in improving the hybrid Dense-UNet201 model.

In the first and second simulations, three different transfer-learning U-nets were investigated for cervical cell segmentation: U-Net has 23.7 million trainable parameters, Efficient-UNetB0 has 10.0 million, Res-UNet50 has 32.5 million, and Dense-UNet201 has 26.1 million.

### 2.7. Performance Evaluation Metrics

The performance metrics for the pap smear image segmentation on validation and test data are reported using loss, accuracy, IoU, and dice coefficient [36].

$$Accuracy = \frac{TP+TN}{TP+TN+FP+FN} \qquad (10)$$

$$Dice\ Coefficient\ = \frac{2\times|P\cap G|}{|P|+|G|} \quad (11)$$

$$IoU = \frac{|P\cap G|}{|P\cup G|} \quad (12)$$

$$loss = 1 - \frac{1}{C}\sum_{c=1}^{C}\frac{2\sum P_c G_c + \epsilon}{\sum P_c + \sum C_c + \epsilon} \quad (13)$$

where $TN$, $FN$, and $TP$ are true negatives, false negatives, true positives, respectively; $FP$ is false positive; $P$ is the ground-truth object, and $G$ is the predicted object. $C$ is number of classes and $\epsilon$ is a small number to avoid division by zero. In this study, the categorical dice loss was used as the loss metric [37].

## 3. RESULTS AND DISCUSSION

This study evaluated the Hybrid Dense-UNet201 method in three scenarios to assess its performance in multi-class semantic segmentation. Simulations were conducted using the SIPaKMeD dataset. The performance of the Hybrid Dense-UNet201 model was compared with that of three semantic segmentation models: U-Net, Res-UNet50, and Efficient-UNetB0.

### 3.1. The first scenario

Figure 5 shows the simulation results for the four semantic segmentation models. The Dense-UNet201 model achieved the lowest loss (15.26%). This indicates a superior reduction in the prediction errors. This was followed closely by Res-UNet50 (17.45%) and Efficient-UNetB0 (18.10%), demonstrating strong generalization. The UNet model exhibited the highest loss (34.86%), suggesting a higher difficulty in learning complex segmentation patterns than the other models. Regarding accuracy, Dense-UNet201 outperformed all other models, achieving an accuracy of 90.54%, followed by Res-UNet50 (89.81%) and Efficient-UNetB0 (88.56%). The standard UNet model exhibited the lowest accuracy (78.61%).

The IoU showed that Dense-UNet201 performed the best (78.12%). Res-UNet50 (74.85%) and Efficient-UNetB0 (74.78%) followed closely. However, the UNet model achieved the lowest IoU (56.31%) among all models. The Dice coefficient results aligned with the IoU trends, where Dense-UNet201 achieved the highest Dice coefficient (86.44%), followed by Res-UNet50 (84.43%) and Efficient-UNetB0 (83.90%). The standard UNet model recorded this study's lowest Dice coefficient (69.39%). The simulation results demonstrated that replacing the standard U-Net encoder with a CNN architecture improved the performance of the U-Net model. Additionally, incorporating transfer learning into the CNN architecture used as the encoder further enhances the segmentation performance for Pap smear images. Overall, the first simulation indicated that Dense-UNet201 was the most effective model for single-cell segmentation in pap smear images, outperforming the standard U-Net, Res-UNet50, and Efficient-UNetB0 models.

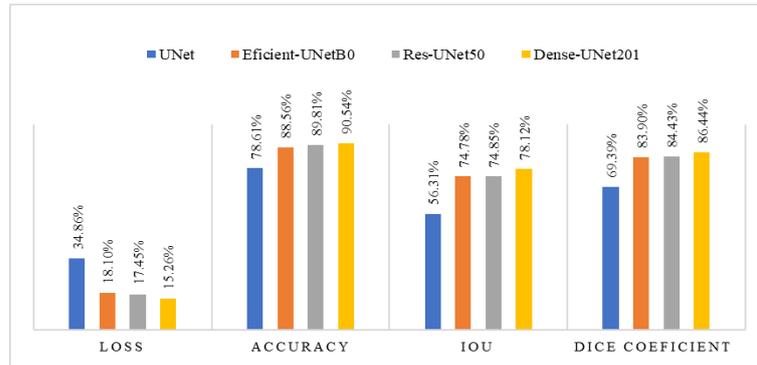

Figure 5. The simulation results of the four semantic segmentation models were evaluated using the SIPaKMeD dataset.

### 3.2. The second scenario

The results of the second scenario are shown in Figure 6. The simulation results of the four semantic segmentation models were evaluated using the SIPaKMeD dataset, which was enhanced using SMO hybrid PMD-CLAHE. The standard U-Net model, with a loss of 33.97% and an accuracy of 79.27%, showed moderate performance in segmentation, as indicated by an IoU of 57.42% and a Dice coefficient of 70.35%. In contrast, the Efficient-UNetB0 model, which utilizes a more lightweight design, significantly reduced the loss to 19.57% and increased the accuracy to 88.45%, resulting in an IoU of 72.50% and a dice coefficient of 82.42%. This indicates a significant improvement in the segmentation performance after the dataset was improved using SMO Hybrid PMD-CLAHE. The Res-UNet50 model further enhanced the performance, with a loss of 17.89%, accuracy of 89.69%, IoU of 74.07%, and Dice coefficient of 83.96%. The inclusion of residual connections in Res-UNet50 helps improve the gradient flow. This resulted in a slight yet significant performance increase over Efficient-UNetB0. Finally, the Dense-UNet201 model outperformed all other models, achieving the lowest loss (13.28%), highest accuracy (92.02%), IoU (80.76%), and Dice coefficient (88.18%). Pap smear image quality enhancement using a hybrid PMD-CLAHE optimized with SMO had a positive impact, significantly improving the performance of the selected segmentation model. The Dense-UNet201 performance improved, reducing the loss value by 0.7% while increasing the IoU by 1.49%, the Dice coefficient by 2.63%, and accuracy by 1.74%.

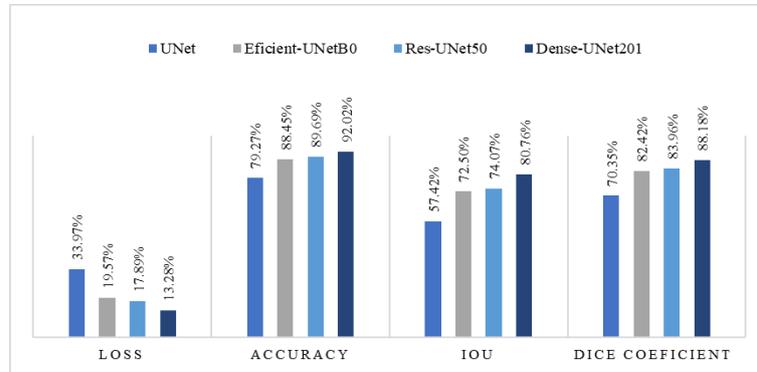

Figure 6. The simulation results of the four semantic segmentation models were evaluated using the SIPaKMeD dataset, which was enhanced using SMO hybrid PMD-CLAHE.

### 3.3. The third scenario

In this scenario, Dense-UNet201 was optimized using SMO and evaluated using the SIPaKMeD dataset, which was enhanced using SMO hybrid PMD-CLAHE. The simulation results and comparison with Dense-UNet201 without optimization are shown in Figure 7. For the Dense-UNet201 model, the loss, accuracy, IoU, and dice coefficient were 13.3%, 92.0%, 80.8%, and 88.2%, respectively. These values reflect a strong performance of Dense-UNet201. Its decent intersection over union (IoU) and dice coefficient scores demonstrated the model's ability to differentiate key features.

A substantial performance boost was observed when SMO optimization was applied to Dense-UNet201. SMO significantly enhanced the Dense-UNet201 performance in segmenting pap smear images. The optimized model showed a substantial reduction in loss, decreasing from 13.28% to 5.03%. This indicates a more stable training process and improved error minimization. Additionally, the segmentation accuracy improved from 92.02% to 96.16%, highlighting the model's enhanced ability to identify and classify cellular structures accurately. The Intersection over Union (IoU) score increased from 80.76% to 91.63%, reflecting a higher overlap between the segmented output and ground truth. Furthermore, the Dice coefficient, which measures segmentation quality, increased from 88.18% to 95.63%, confirming greater precision and consistency in cell segmentation. These results demonstrate that SMO optimization positively impacts Dense-UNet201.

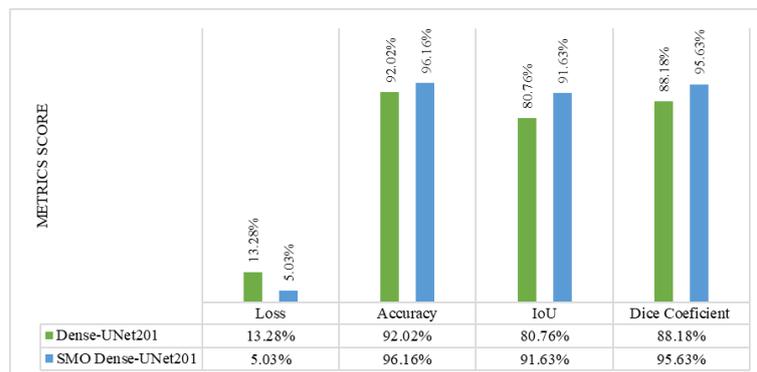

Figure 7. The simulation results of the proposed models (SMO Dense-UNet201) were evaluated using the SIPaKMeD dataset, which was enhanced using the SMO hybrid PMD-CLAHE.

### 4. CONCLUSION

This study evaluated the Hybrid Dense-UNet201 model in three scenarios for multi-class semantic segmentation of Pap smear images. Results showed that Hybrid Dense-UNet201 consistently outperformed U-Net, Res-UNet50, and Efficient-UNetB0 in accuracy, IoU, and Dice coefficient. In the first scenario, Hybrid Dense-UNet201 achieved an accuracy of 90.54%, IoU of 78.12%, and Dice coefficient of 86.44%, with the lowest loss (15.26%). The second scenario, which applied the SMO hybrid PMD-CLAHE preprocessing, significantly improved segmentation across all models. Hybrid Dense-UNet201 further enhanced its performance, achieving 92.02% accuracy, 80.76% IoU, and 88.18% Dice coefficient, underscoring the importance of preprocessing. Optimizing Hybrid Dense-UNet201 with SMO in the third scenario led to remarkable improvements, with accuracy, IoU, and Dice coefficient reaching 96.2%, 91.6%, and 95.6%, respectively. These findings demonstrate that SMO Dense-UNet201 is a highly effective Pap smear image segmentation approach. Future research should explore its application to other medical imaging tasks and investigate additional optimization techniques for further improvements.

**FUNDING INFORMATION**

This research was funded by the Ministry of Higher Education, Science, and Technology of the Republic of Indonesia and the Indonesian Education Foundation (LPDP) through the Center for Higher Education Funding and Assessment (PPAPT) under the Indonesian Education Scholarship (BPI) program.

**CONFLICT OF INTEREST STATEMENT**

The authors confirm that they have no conflicts of interest regarding the publication of this work and no financial or personal affiliations that could affect its content.

**DATA AVAILABILITY**

The publicly available SIPaKMeD dataset was used in this study and can be accessed at https://bit.ly/SIPaKMeD for further research in medical image processing and cervical cancer classification.